\documentclass[11pt]{article} %!! ARXIV
\usepackage{etoolbox} % ! Insight Journal

\newtoggle{arxiv}
\toggletrue{arxiv} % !! ARXIV
\usepackage[utf8]{inputenc}
\usepackage[dvips]{graphicx}

\usepackage{amsmath,amsfonts,amssymb,amsthm}

\usepackage[%dvips,
bookmarks,
bookmarksopen,
backref,
colorlinks,linkcolor={blue},citecolor={blue},urlcolor={blue}
]{hyperref}

\iftoggle{arxiv}{
\voffset=-2cm
\hoffset=-1.5cm
\textwidth=16.5cm
\textheight=23cm
}{}

\def\cE{\mathcal E}

\def\mR{\mathbb R}

\DeclareMathOperator\Id{Id}

\def\R{\mR}
\def\<{\langle}
\def\>{\rangle}

\def\bn{{\bold n}}

\DeclareMathOperator\diver{div}
\DeclareMathOperator\Diff{\mathbf{D}} %diffusion tensor
\DeclareMathOperator\Struct{\mathbf{S}} %Structure tensor

\usepackage{enumerate}

\iftoggle{arxiv}{
\newcommand{\doxygen}[1]{ {\bf #1} }
}{}
\title{Anisotropic Diffusion in ITK}

\iftoggle{arxiv}{}{
\newcommand{\IJhandlerIDnumber}{1338}

\release{0.00}
}

\iftoggle{arxiv}{
\author{Jean-Marie Mirebeau$^1$, Jérôme Fehrenbach$^2$, Laurent Risser$^2$, Shaza Tobji$^2$} 
}{
\author{Jean-Marie Mirebeau$^1$, Jérôme Fehrenbach$^2$, Laurent Risser$^2$, Shaza Tobji$^2$} 
\authoraddress{%
$^{1}$University Paris-Dauphine, Laboratory Ceremade, CNRS\\
$^{2}$Institut de Math\'ematiques de Toulouse, Universit\'e Paul Sabatier, CNRS\\
}
} %iftoggle
\begin{document}

%
% Add hyperlink to the web location and license of the paper.
% The argument of this command is the handler identifier given
% by the Insight Journal to this paper.
% 
\iftoggle{arxiv}{}{
\IJhandlefooter{\IJhandlerIDnumber}
}

\ifpdf
\else
   %
   % Commands for including Graphics when using latex
   % 
   \DeclareGraphicsExtensions{.eps,.jpg,.gif,.tiff,.bmp,.png}
   \DeclareGraphicsRule{.jpg}{eps}{.jpg.bb}{`convert #1 eps:-}
   \DeclareGraphicsRule{.gif}{eps}{.gif.bb}{`convert #1 eps:-}
   \DeclareGraphicsRule{.tiff}{eps}{.tiff.bb}{`convert #1 eps:-}
   \DeclareGraphicsRule{.bmp}{eps}{.bmp.bb}{`convert #1 eps:-}
   \DeclareGraphicsRule{.png}{eps}{.png.bb}{`convert #1 eps:-}
\fi

\maketitle

\iftoggle{arxiv}{}{
\ifhtml
\chapter*{Front Matter\label{front}}
\fi
}

\begin{abstract}
\noindent
Anisotropic Non-Linear Diffusion is a powerful image processing technique, which allows to simultaneously  remove the noise and enhance  sharp features in two or three dimensional images. 
Anisotropic Diffusion is understood here in the sense of Weickert, meaning that  diffusion tensors are anisotropic and reflect the local orientation of  image features. This is in contrast with the non-linear diffusion filter of Perona and Malik, which only involves scalar diffusion coefficients, in other words \emph{isotropic} diffusion tensors. 
In this paper, we present an anisotropic non-linear diffusion technique we implemented in ITK. This technique is based on a recent adaptive scheme making the diffusion stable and requiring limited numerical resources. 
%implement anisotropic diffusion in ITK using a recent adaptive scheme, which is stable and has a limited numerical cost. 
\end{abstract}

\iftoggle{arxiv}{}{
\IJhandlenote{\IJhandlerIDnumber}
}

\tableofcontents

%\section{Non-linear Anisotropic Diffusion in image processing}
%\label{sec:Math}
%\subsection{Mathematical background}

\footnotetext[1]{University Paris-Dauphine, Laboratory Ceremade, CNRS.}
\footnotetext[2]{Institut de Math\'ematiques de Toulouse, Universit\'e Paul Sabatier, CNRS.\\
Young researcher ANR program: Numerical Schemes using Lattice Basis Reduction. NS-LBR ANR-13-JS01-0003-01}
\section{Introduction}

A digital image is usually a large two or three dimensional array of pixel values (typically scalars or vectors).
Image processing methods based on Partial Differential Equations (PDEs) regard images as approximations of continuous objects, namely functions from an image domain to a pixel space, to which physics-inspired evolution rules can be applied. Among them, Non-Linear Anisotropic Diffusion (NLAD) is a variant of the heat equation, generalized in two regards: \emph{Non-Linearity} and \emph{Anisotropy}. 

\emph{Anisotropy} in diffusion means that the smoothing induced by the PDE can be favored in some directions and prevented in others. This is specified by local eigenvectors and eigenvalues of the diffusion tensor field  (see \S\ref{sec:LAD}). 
Diffusion coefficients are thus location \emph{and direction} dependent, generalizing the approach of Perona and Malik \cite{Perona:1990ej} which is only location dependent.  
Importantly, efficient schemes for anisotropic diffusion have been recently made possible by the breakthrough in \cite{Fehrenbach:2013ut}. This has motivated the development of the ITK module presented in this paper.

\begin{figure}
\includegraphics[width=4cm]{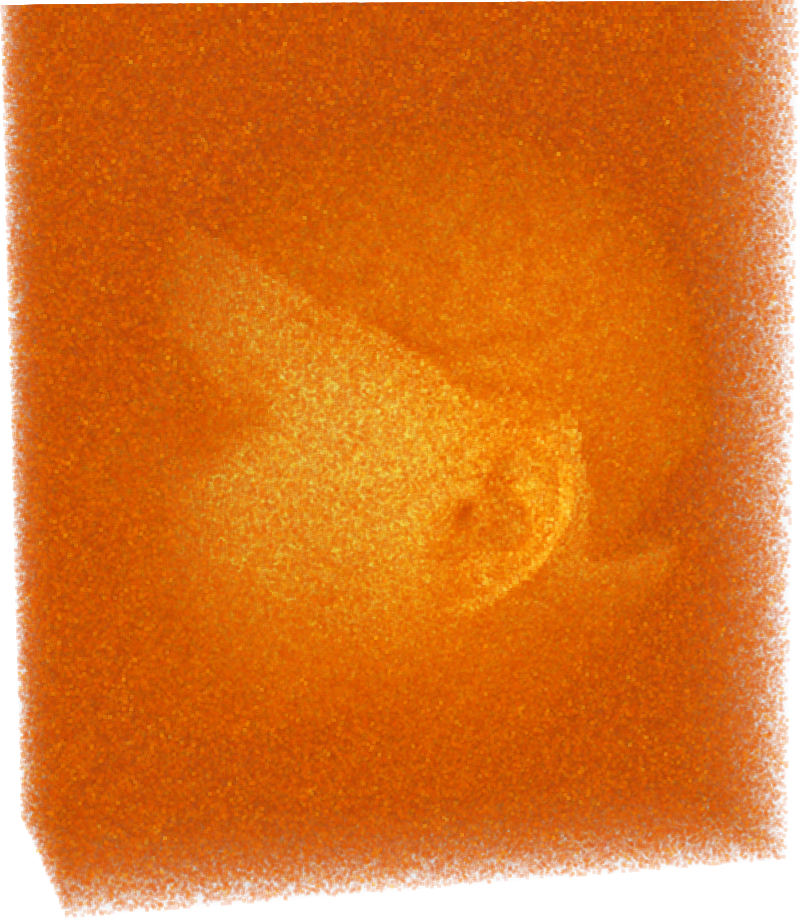}
\includegraphics[width=4cm]{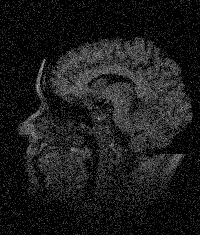}
\includegraphics[width=4cm]{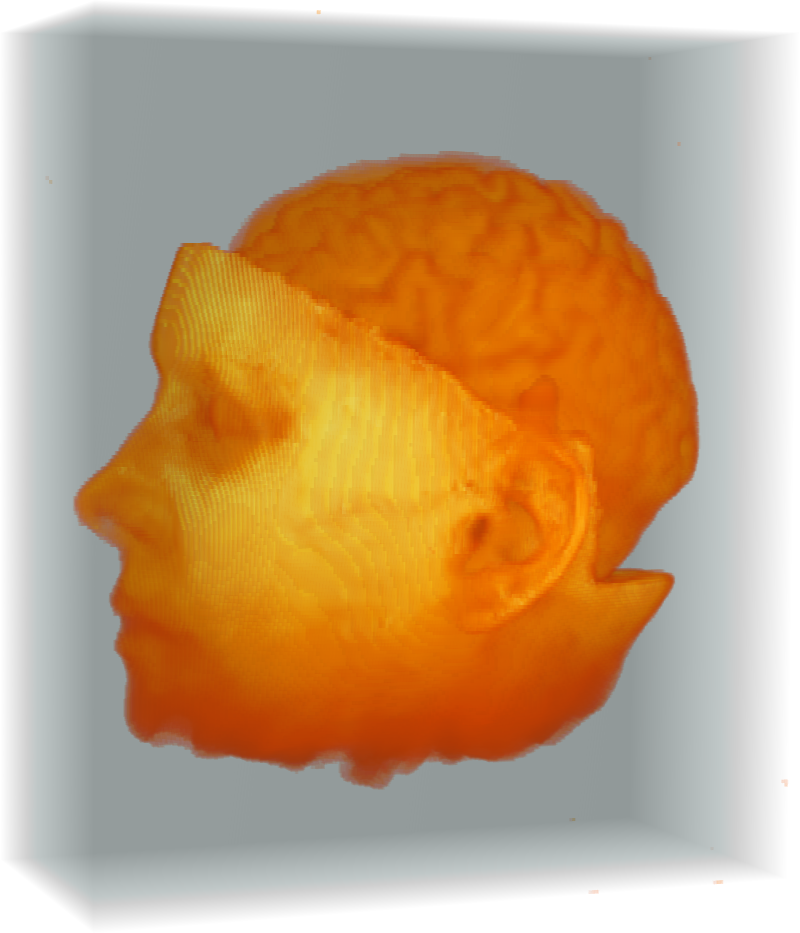}
\includegraphics[width=4cm]{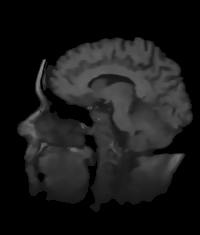}
%./itkDiffusion ../Images/mrbrain_noisy_01.hdf5 mrbrain_smoothed_01.hdf5 5 0.003 Edge #Good
\caption{Left: Volume plot and slice of an FMRI of the human skull, with artificially added Gaussian noise of variance $0.01$ (data range: $[0,1]$). In the volume plot, values below $0.08$ are shown transparent. Right: Effect of cEED. Parameters: $T=5$, $\lambda=0.003$.}
\label{fig:Skull}
\end{figure}

\begin{figure}
\includegraphics[width=4cm]{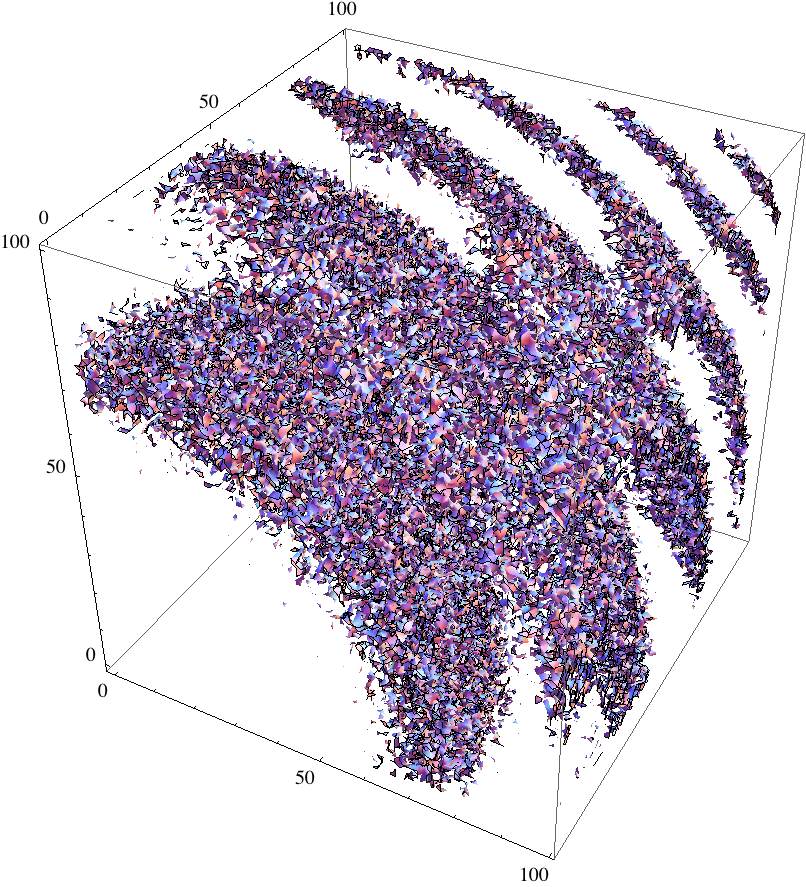}
\includegraphics[width=4cm]{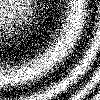}
\includegraphics[width=4cm]{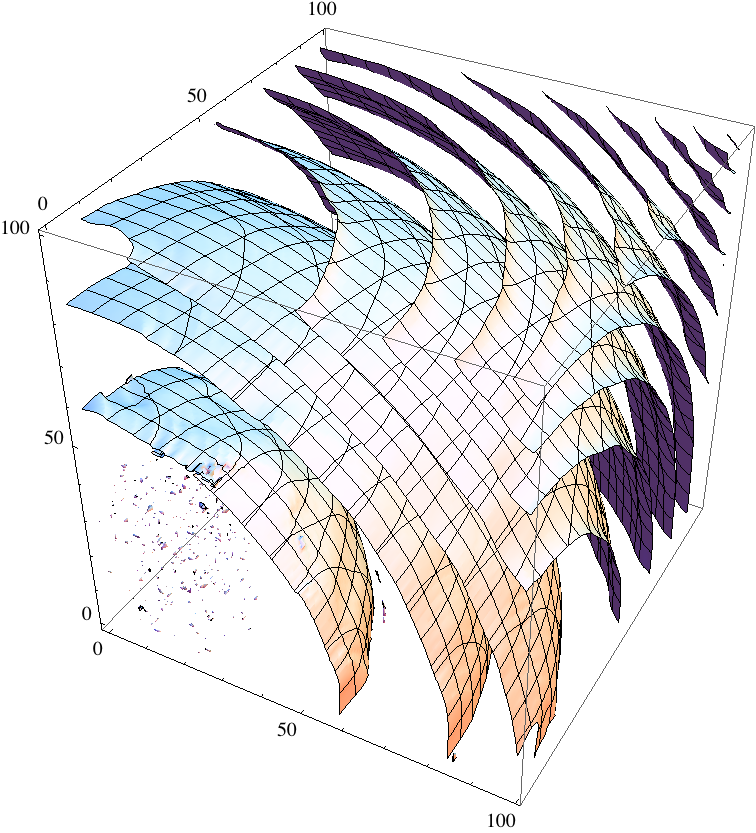}
\includegraphics[width=4cm]{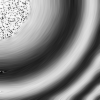}
%./itkDiffusion ../Images/EdgeEnhancingDiffusion3D_Noisy.hdf5 EdgeEnhancingDiffusion3D_smoothed.hdf5 10 0.02 CoherenceCertified 4 10 
\caption{Synthetic function $\cos(\|x\|^3)$, $x \in [0,1]^3$, corrupted with Gaussian noise of variance $0.5$. Left: level lines, and slice $\{x=0.1\}$. Right: effect of CED. By construction, CED barely affects the neighborhood of the origin $(0,0,0)$, where no specific coherent direction can be determined. Parameters: $T=10$, $\lambda=0.02$, $\sigma=4$, $\rho=10$.
}
\label{fig:Cos3D}
\end{figure}

\begin{figure}
\includegraphics[width=4cm]{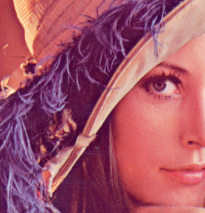}
\includegraphics[width=4cm]{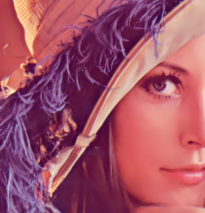}
\includegraphics[width=4cm]{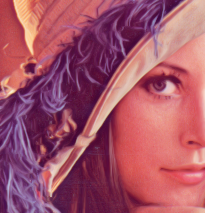}
\includegraphics[width=4cm]{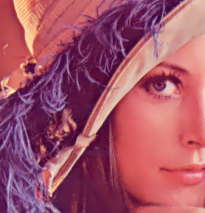}
%2 0.003 cCED 0.5 2 4
\caption{I: Detail of the Lena image. II: cEED removes noise and preserves most image detail. III: cCED enhances the flow structure of the hat plumes, without much affecting the rest of the image. In particular it does not remove noise. IV: Isotropic diffusion (\ref{eqdef:Isotropic}, Perona Malik like) removes noise in the interior of the image regions, but leaves some noise close to image contours such as the border of Lena's cheek.
Parameters: $T=2$, $\lambda = 0.003$, $m=4$, others default.
% , on a short time $T=5$, enhances the flow structure of the hat plumes without much affecting the rest of the image. III, IV: effect of long time diffusion $T=80$ with (III) cEED  and (IV) Isotropic (\ref{eqdef:Isotropic}, Perona-Malik like)  diffusion. Both produce large regions where the filtered image is constant. The contours of these regions are smoother with cEED, see e.g. the hat.
}
\label{fig:Lena}
\end{figure}

\begin{figure}
\includegraphics[width=4cm]{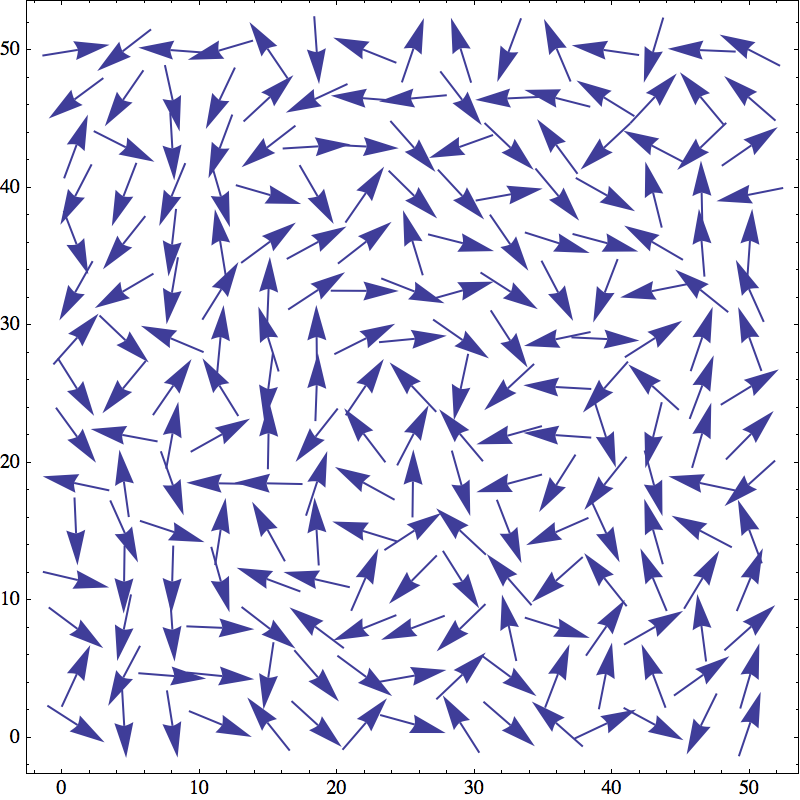}
\includegraphics[width=4cm]{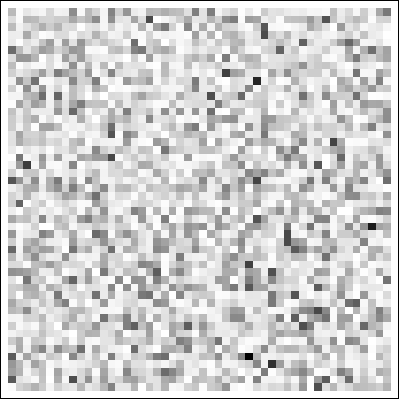}
\includegraphics[width=4cm]{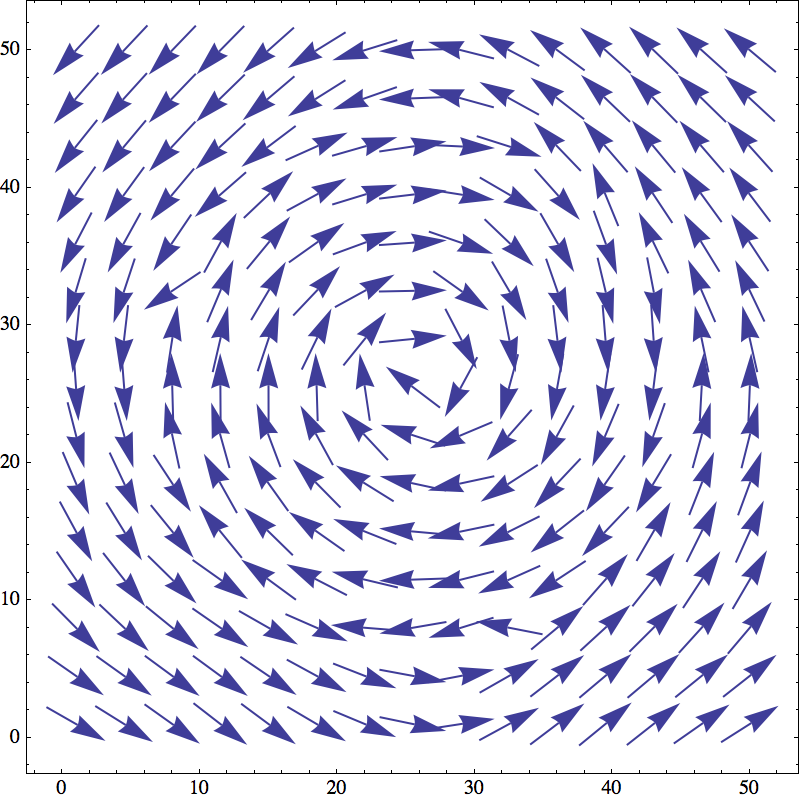}
\includegraphics[width=4cm]{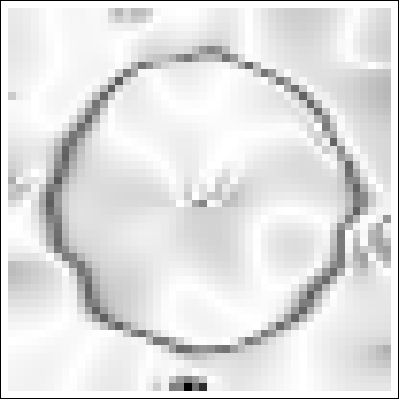}
%10 0.05 cEED 
\caption{Left: Directions and norms (black=0, white=1) of the vector field $v(x) := {\rm sign}(\|x\|-1)\ x^\perp/\|x\|$ on $[-1.3,1.3]^2$, degraded by gaussian noise of variance $2$, on a $50 \times 50$ grid. Norms of these vectors (black=0, white=1).   
%\textcolor{red}{($v(x) := {\rm sign}(\|x\|-1)\ x^\perp/(\|x \|+\epsilon)$, with $\epsilon$ a small value (?) -- je peux essayer de regarder si j'arrive a generer une image de vecteurs a partir de tes champs pour rendre l'aspect direction si tu veux)}
Right: Effect of cEED. Note that the streamlines are better reconstructed, and that the norms vanish along the (approximate) circle $\|x\|=1$ where $v(x)$ changes sign so that there is a cancellation effect (and likewise close to the vector field singularity at the center).
Parameters: $T=10$, $\lambda=0.05$, others default.
%Vector field denoised by linear anisotropic diffusion (LAD), with a well chosen anisotropic diffusion tensor field, which prevents smoothing accross the circle $\{\|x\|=1\}$ where the vector field $v$ has a discontinuity. Norms of these vectors. Note that these norms are close to $1 = \|v(x)\|$, except on a thin band along 
%(* No : cEED. $T=10$, $\lambda=0.05$. *)
}
\label{fig:Vector}
\end{figure}

\emph{Non-Linearity} in diffusion means that  diffusion tensors are automatically generated from the processed image. We implemented the strategies of Weickert \cite{weickert1998anisotropic} and we give a simple framework for designing extensions and variants, see \S\ref{sec:NLAD}. 
The implemented filters and their parameters are described  in \S \ref{sec:ImplementationParameters}. 
Figures \ref{fig:PacMan} and \ref{fig:FingerPrint} illustrate their effect on 2D images; Figures \ref{fig:Skull} and \ref{fig:Cos3D} on 3D images; Figures \ref{fig:Lena} and \ref{fig:Vector} on color and vector images.

%Images of dimension two and three are supported, with pixels of scalar, vector and color type.
A possible application of NLAD is to enhance a fingerprint image by smoothing tangentially to the lines.
Evidence is also plentiful for NLAD relevance in many other image processing applications, but its use has been limited by technical aspects so far. We intend to alleviate such limitations with the present contribution.

\paragraph{Notations:} 
Let $d \in \{2,3\}$ denote the image dimension, let $\Omega \subset \mR^d$ be the image domain, and let $V$ be the pixel space (e.g.\ $[0,1]$ for grayscale, $[0,1]^3$ for color, $\mR^d$ for vectors). 
Throughout the paper, we informally consider an idealized \emph{cartoon} image model, involving a set $\Gamma \subset \Omega$ of \emph{image contours} of dimension $d-1$. The processed image $u : \Omega \to V$ is smooth on $\Omega \setminus \Gamma$, but has discontinuities across $\Gamma$, and is overall corrupted by e.g.\ by additive white noise. A key feature of NLAD is its ability to detect the set $\Gamma$ and smoothen tangentially to it.
%A key feature of NLAD is the ability to smooth
%
%In practice the proposed filters also work well with more complex images, which may e.g.\ have oscillating textures see Figure \ref{fig:FingerPrint}. The image $u$ may also be corrupted, e.g.\ by additive white noise.
%
%
% Among PDEs, Non-Linear Anisotropic Diffusion (NLAD) has shown its relevance for image processing. 
% NLAD can, when properly tuned, simultaneously remove the noise present in an image, and enhance its sharp features. 
%The scalar and vector pixel types are supported, as well as color images.
%Non-Linear Anisotropic Diffusion is a variant of the heat equation, generalized in two regards. First, diffusion is Anisotropic in other words facilitated in some directions and prevented in others, see \S \ref{sec:LAD}. Second, diffusion is Non-Linear, meaning that the diffusion directions and coefficients are defined in terms of the processed image, see \S \ref{sec:NLAD}. After these two mathematical sections, we describe the implemented filters, and their parameters, see \S \ref{sec:ImplementationParameters}.
 %These two aspects of the PDE, and their relevance for image processing, are discussed \S \ref{sec:LAD} and \ref{sec:NLAD}. Applications examples, and indications on parameter settings, are presented in \S \ref{sec:Examples}.
%
%We denote by $d \in \{2,3\}$ the image dimension, by $\Omega \subset \R^d$ the image domain, and by $V$ the pixel space (e.g.\ $[0,1]$ for grayscale, $[0,1]^3$ for color, $\mR^d$ for vectors). 
%
Finally, let $S_d^+$ denote the collection of symmetric positive definite $d \times d$ matrices, and let $\Id$ be the identity matrix. 
To each $D \in S_d^+$ we associate the norm $\|e\|_D := \sqrt{\<e, D e\>}$, $e \in \R^d$, where $\<\cdot, \cdot\>$ denotes the standard scalar product on $\R^d$.

\section{Linear Anisotropic diffusion}
\label{sec:LAD}
Linear Anisotropic Diffusion%
\footnote{%
We use here the terminology of Weickert \cite{weickert1998anisotropic}. Perona-Malik diffusion, which uses an adaptive scalar tensor field similar to \eqref{eq:PeronaMalikTensors}, is in contrast a \emph{Non-Linear Isotropic} Diffusion equation. 
} (LAD),
in divergence form, is an elliptic PDE which reads
\begin{equation}
\label{eq:LinearAnisotropicDiffusion}
\partial_t u = \diver(\Diff \nabla u),
\end{equation}
where $\Diff : \Omega \to S_d^+$ is a given field of symmetric positive definite diffusion tensors. Eigenvectors of these tensors define preferential diffusion directions, and the eigenvalues their corresponding coefficients.
Evolution rule \eqref{eq:LinearAnisotropicDiffusion} is complemented with an initial condition $u(0, \cdot) = u_0$ at time $t=0$. If $u_0$ has pixels of vector type, then their components are treated independently.
We use Neumann%
\footnote{%
Neuman boundary conditions must take into account the geometry defined by the diffusion tensor field. They take the form 
$\< \nabla u(x), \Diff(x) \bn(x)\> = 0$, where $\bn(x)$ denotes the unit outward normal at $x \in \partial \Omega$.
}
conditions on the domain boundary $\partial \Omega$, as is common in image processing. LAD is formally a continuous gradient descent for the elliptic energy 
\begin{equation}
\label{eq:AnisotropicEnergy}
\cE(u)
%\|\nabla u\|_{\Diff}^2 
:= \int_\Omega \|\nabla u(x)\|^2_{\Diff(x)} dx.
\end{equation}
Qualitative effects of LAD strongly depend on the chosen field $\Diff$ of diffusion tensors. Choosing $\Diff = \Id$ identically on $\Omega$ yields the standard heat equation, which qualitative properties are well known in image analysis: any noise present in the image $u$ is quickly eliminated, but in the meanwhile all  image sharp features are blurred. 
%, and therefore makes no distinction between  
%Indeed both image noise and image contours yield strong gradients, contributing to the energy \eqref{eq:AnisotropicDiffusion} 
%Indeed, tends to diffusion eliminate by smoothing the strong image 
%The reason is that diffusion tends to smooth regions which have  
%The reason is that the strong image gradients localized around  the processed image sharp features, contribute 

\begin{figure}
\includegraphics[width=4cm]{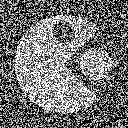}
\includegraphics[width=4cm]{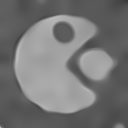}
\includegraphics[width=4cm]{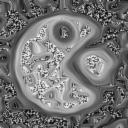}
\includegraphics[width=4cm]{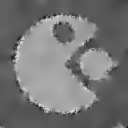}

%./itkDiffusion ../Images/PacMan.png PacMan_cEED.png 20 0.05 cEED 3
%./itkDiffusion ../Images/PacMan.png PacMan_cCED.png 20 0.05 cCED 3
%./itkDiffusion ../Images/PacMan.png PacMan_I.png 20 0.05 Isotropic 3

\caption{I: Source image. II: cEED is the best pick.  III: cCED only affects the directionally coherent parts, the image region contours, which is clearly inadequate.
IV: Isotropic (\ref{eqdef:Isotropic}, Perona-Malik like) diffusion leaves some noise trapped along the image contours. Parameters: $T=20$, $\lambda=0.05$, $\sigma=3$, others default.}
\label{fig:PacMan}
\end{figure}

This undesirable side effect can be limited with a proper choice of diffusion tensors $\Diff : \Omega \to S_d^+$. 
Indeed LAD smoothes primarily the image features which contribute strongly to the energy \eqref{eq:AnisotropicEnergy}. 
%Assume for instance that the processed image is a photography of a collection of objects, and denote by $\Gamma \subset \Omega$ the collection of curves delimiting the objects%
%\footnote{%
%In practice, NLAD does not require any knowledge of the image object contours, but may actually help detect them.
%}.
In the spirit of Perona and Malik, one can introduce an \emph{isotropic} but \emph{variable} conductivity $\Diff(x) = c(x) \Id$, with $c(x) \ll 1$ for all $x$ close to the image contours $\Gamma$, see introduction. 
Smoothing is prevented in the neighborhood of $\Gamma$, which preserves the contours sharpness, but also traps some noise along them, see Figure \ref{fig:PacMan} (IV). % image ??
A more elaborate approach is to construct \emph{anisotropic} diffusion tensors $\Diff(x)$, which favor diffusion \emph{tangentially} to the contours curves $\Gamma$, but simultaneously prevent diffusion transversally to these curves and between different image regions. All noise is eliminated, yet image discontinuities are preserved, see Figure \ref{fig:PacMan} (II).

Numerical schemes for LAD are in general non-trivial due to interaction between the anisotropic geometry of the diffusion tensors, and the  cartesian structure of the pixel grid. 
The authors recently developed \cite{Fehrenbach:2013ut} a numerical scheme which handles this interaction using special tools from discrete geometry, named Lattice Basis Reduction (LBR). It provides strong mathematical guarantees (consistency, stability, maximum-principle) for a limited numerical cost.

\section{Non-Linear Anisotropic Diffusion.}
\label{sec:NLAD}
\begin{figure}
\includegraphics[width=4cm]{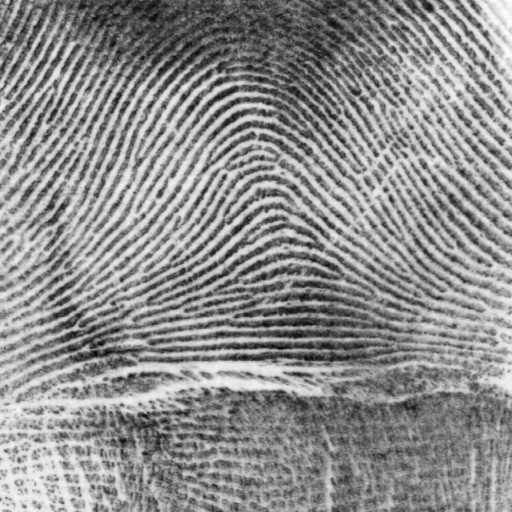}
\includegraphics[width=4cm]{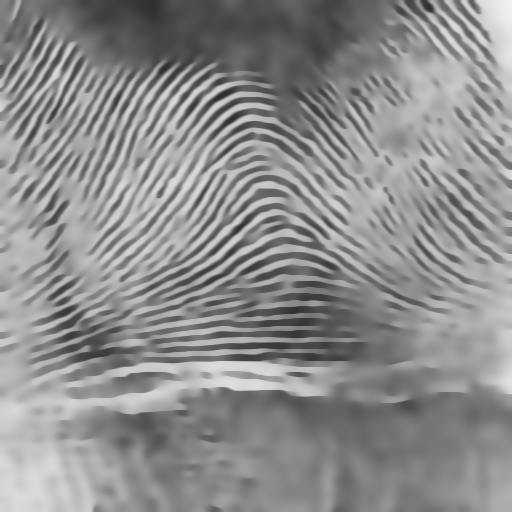}
\includegraphics[width=4cm]{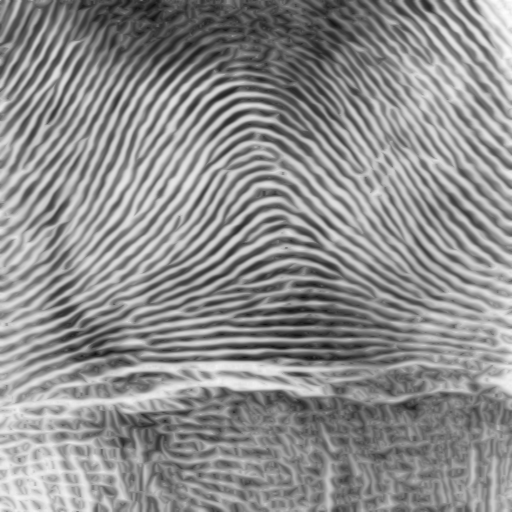}
\includegraphics[width=4cm]{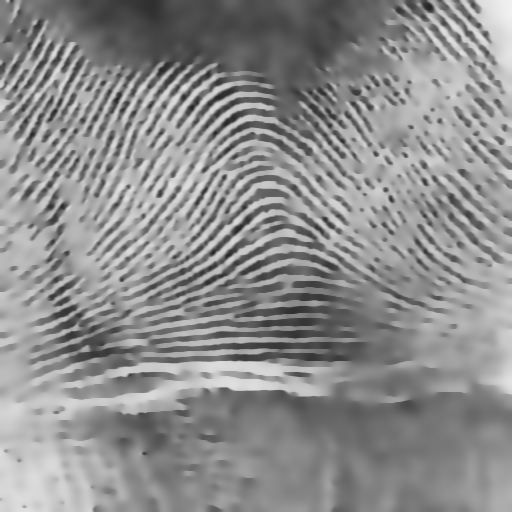}
%./itkDiffusion ../Images/FingerPrint.png FingerPrint_smoothed_20_E.png 20 0.02 Edge
%./itkDiffusion ../Images/FingerPrint.png FingerPrint_smoothed_20_C.png 20 0.02 Coherence
%./itkDiffusion ../Images/FingerPrint.png FingerPrint_smoothed_20_I.png 20 0.02 Isotropic
\caption{I: Source image. II: cEED is not advised, since it blurs the junctions of the fingerprint lines. III: cCED is the best pick, since it enhances the fingerprint lines, and does not blur the more complex regions. IV: Isotropic (\ref{eqdef:Isotropic}, Perona-Malik like) diffusion either fails to remove noise, or blurs the fingerprint lines, depending on the image region.
Parameters: $T=20$, $\lambda=0.02$, others default.}
\label{fig:FingerPrint}
\end{figure}

Linear Anisotropic Diffusion, discussed in \S\ref{sec:LAD}, requires two main inputs: an image $u_0$ serving as an initial condition, and a field of diffusion tensors. In order to reduce user input, the diffusion tensors $\Diff$  can be defined in terms of the filtered image $u$. The resulting PDE is called non-linear anisotropic diffusion 
\begin{equation}
\label{eq:NLAD}
%\partial_t u(x,t) = \diver(\Diff(x; u) \nabla u(x,t)).
\partial_t u = \diver(\Diff_u \nabla u),
\end{equation}
complemented, again, with Neumann boundary conditions. 
Perona and Malik \cite{Perona:1990ej} suggested to use the following \emph{non-linear isotropic} (i.e.\ proportional to the identity matrix) tensors
\begin{equation}
\label{eq:PeronaMalikTensors}
\Diff_u(x,t) := c_u (x,t) \Id, \qquad \text{with} \quad c_u(x,t) := \frac 1 {\sqrt{1+ \|\nabla u(x,t)\|^2/\lambda^2}},
\end{equation}
where $\lambda>0$ is a user specified constant. Diffusion is prevented where the conductivity $c_u(x,t)$ is small, in other words where $\|\nabla u(x,t)\|$ is large, such as along the image contours $\Gamma$. Perona-Malik diffusion is already available%
\footnote{%
Under the name (misleading with our conventions) \doxygen{GradientAnisotropicDiffusionImageFilter}.
}
in ITK. It has been the subject of considerable academic and industrial interest revealing that, in spite of its numerous qualities, it is mathematically ill posed, unstable, often leads to unsightly ``staircasing'' visual artifacts, and is not adequate for oscillating patterns as in Figure \ref{fig:FingerPrint}. 

We describe in the following Coherence Enhancing Diffusion (CED) and Edge Enhancing Diffusion (EED), which are based on more complex tensor constructions introduced by Weickert \cite{weickert1998anisotropic}. Our first ingredient is the Gaussian convolution kernel: given a standard deviation $\sigma > 0$
\begin{equation}
K_\sigma (x)  := \frac 1 {\sigma^{d}} K_1\left(\frac x \sigma\right), \qquad \text{where} \quad  K_1(x):= \frac 1 {(2 \pi)^{d}} \exp\left(\frac{-\|x\|^2} 2\right).
\end{equation}
The structure tensor $\Struct_u : \Omega \to S_d^+$, defined below, is a robust estimator of the gradient direction in an image $u$, even if this image has oscillating textures. It depends on two small positive parameters: the noise scale $\sigma$, and the feature scale $\rho$. We denote by $*$ the convolution operator, and by $v \otimes v = v v^{\rm T}$ the self outer product, which yields a semi-definite symmetric matrix.
\begin{equation}
\label{eq:StructureTensor}
\Struct_u := K_\rho * (\nabla u_\sigma \otimes \nabla u_\sigma), \qquad \text{where} \quad u_\sigma := K_\sigma * u.
\end{equation}
If $u$ is an image with vector pixels, then $\Struct_u$ is the \emph{sum} of the structure tensors associated to the components of $u$.
Assume that $u$ is a scalar image, fix a time $t$ and a point $x \in \Omega$, and denote $S := \Struct_u(x,t)$, $v := \nabla u(x,t)$. Let also $\lambda_1 \leq \cdots \leq \lambda_d$ denote the eigenvalues of $S$, sorted by increasing magnitude, and $e_1, \cdots, e_d$ the corresponding unit eigenvectors. If $u$ is sufficiently smooth, then the largest eigenvalue approximates the gradient squared norm: $\lambda_d \approx \|v\|^2$, while the corresponding eigenvector approximates the gradient direction: $e_d \approx \pm \frac v {\|v\|}$.

Weickert's diffusion tensors $D := \Diff_u(x,t)$, are defined in terms of this eigen-analysis of the structure tensor $S := \Struct_u(x,t)$:
\begin{equation}
\label{eq:StructToDiff}
D = \sum_{1 \leq i \leq d} \mu_i e_i \otimes e_i, \quad \text{ where } \quad S = \sum_{1 \leq i \leq d} \lambda_i e_i \otimes e_i.
\end{equation}
Smoothing is promoted in the direction $e_i$ if $\mu_i$ is large, and prevented if $\mu_i$ is small, for any $1 \leq i \leq d$.
%Diffusion tensor eigenvalues $\mu_i$, $1 \leq i \leq d$, promote smoothing in the direction $e_i$ when they are large, and prevent it when they are small. 
Weickert's classical constructions are presented in \eqref{eqdef:EED} and \eqref{eqdef:CED}. One should not shy away of designing more complex and application dependent variants; for instance one may want to enhance filaments and tubular structures in 3D data. Three very simple variants \eqref{eqdef:cEED}, \eqref{eqdef:Isotropic} and \eqref{eqdef:cCED} are presented for illustration. All depend on three parameters $\lambda, m, \alpha$. The main one, $\lambda>0$, is an edge detection threshold. The exponent $m$ is typically $2$ or $4$. The small parameter $\alpha$, typically $1/100$, determines the condition number of the diffusion tensors.

\begin{figure}
\centering
\includegraphics[height=4cm]{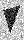}
\includegraphics[height=4cm]{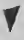}
\includegraphics[height=4cm]{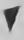}
\hspace{0.3cm}
\includegraphics[height=4cm]{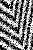}
\includegraphics[height=4cm]{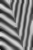}
\includegraphics[height=4cm]{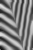}
%./itkDiffusion ../Images/PacMan.png PacMan_smoothed_C_01.png 20 0.001 CoherenceCertified 3
%./itkDiffusion ../Images/Triangle.png Triangle_cEED.png 5 0.03 cEED #Good
%./itkDiffusion ../Images/Triangle.png Triangle_EED.png 5 0.03 EED #Good
%./itkDiffusion ../Images/Oscillations_Noisy1.png Oscillations1_cCED.png 20 0.05 cCED
%./itkDiffusion ../Images/Oscillations_Noisy1.png Oscillations1_CED.png 20 0.05 CED
\caption{
Left: (I) source image, (II) effect of cEED, (III) effect of EED. Note that the triangle corners are better preserved with cEED. Parameters: $T=5$, $\lambda=0.03$.
Right: (I) source image, (II) effect of cCED, (III) effect of CED. Note that cCED better preserves contrast along where the two fronts meet.
Parameters: $T=20$, $\lambda=0.05$.
%(I) cCED with a long time $T=20$, and a low edge detection threshold $\lambda=0.001$, will detect structure in noise (false positive), and enhance it, which can be used for ``artistic effects''. 
%(II) Noisy image of a triangle, (III) effect of cEED, (IV) and EED. Parameters $T=5$, $\lambda=0.03$. Both cEED and EED eliminate noise an preserve sharp edges, but cEED preserves the triangle sharp corners better than EED.
}
\label{fig:Triangle}
\end{figure}

\begin{itemize}
\item
Edge Enhancing Diffusion (EED) aims to avoid significant diffusion across the set $\Gamma$ of image contours, but to allow it anywhere else. Note that for 3D images discontinuity planes will be enhanced, rather than edges.
The first diffusion tensor eigenvalue is $\mu_1 = 1$, because the eigenvector $e_1$ is orthogonal to the image (approximate) gradient direction $e_d$, hence never transverse to $\Gamma$.
Other eigenvalues satisfy $\mu_i \approx \alpha \ll 1$ if $\lambda_i-\lambda_1 \gtrsim \lambda$, and $\mu_i \approx 1$ otherwise. The condition $\lambda_i-\lambda_1 \gtrsim \lambda$ indeed suggests that the eigenvector $e_i$ points through an image contour. Precisely%
\footnote{%
Actually, Weickert uses $\mu_1 = 1$ together with \eqref{eqdef:cEED} for $i>1$. This results in discontinuous diffusion tensors, which is not advisable from a mathematical standpoint, hence the formula \eqref{eqdef:EED}. %However, and in contrast with the presented constructions, this produces a discontinuous field of diffusion tensors, which is not advisable.
}%
: (note that $\mu_1=1$)
\begin{equation}
\label{eqdef:EED}
\mu_i := 1 - (1-\alpha)\exp\left(-\left(\frac \lambda {\lambda_i - \lambda_1}\right)^m\right). 
\end{equation}
The choice of Weickert, to set $\mu_1 = 1$, may lead to undesired effects: one always performs diffusion in at least one direction. An undesirable side effect is that the image is blurred close to the angles of its contour set $\Gamma$. We believe that such salient features should be preserved, hence we introduce a Conservative variant of EED (cEED) for which $\mu_1$ \emph{can} be small, when appropriate, so as to prevent diffusion around the angles of $\Gamma$, see Figure \ref{fig:Triangle} for a comparison. Precisely
\begin{equation}
\label{eqdef:cEED}
\mu_i := 1 - (1-\alpha)\exp\left(-\left(\frac \lambda {\lambda_i}\right)^m\right). 
\end{equation}
If all eigenvalues are set equal $\mu_1=\cdots = \mu_d$, then the diffusion tensors are \emph{isotropic}, in other words scalar multiples of the identity. The following isotropic variant of EED is close in spirit to the Perona-Malik model: diffusion is prevented in the neighborhood of the image contours $\Gamma$, regardless of direction. This construction is implemented purely for comparison with the anisotropic ones, and does not take advantage of the innovative numerical scheme developed by the authors
\begin{equation}
\label{eqdef:Isotropic}
\mu_i := 1 - (1-\alpha)\exp\left(-\left(\frac \lambda {\lambda_d}\right)^m\right). 
\end{equation}
%
%\footnote{%
%Note the minor difference with \cite{weickert1998anisotropic}, which imposes $\mu_1=1$. 
%Note that \cite{weickert1998anisotropic} imposes $\mu_1=1$, but this results in discontinuous diffusion tensors, and tends to artificially smooth edge corners.
%}

\item
Coherence Enhancing Diffusion (CED) prevents diffusion except along local image structures which have a coherent direction.
The diffusion tensor eigenvalues satisfy $\mu_i \approx \alpha \ll 1$, \emph{unless} if $\lambda_d - \lambda_i \gtrsim \lambda$ in which case $\mu_i \approx  1$. The condition $\lambda_d - \lambda_i \gtrsim \lambda$ indeed suggests that $e_d$ points through an image feature, and that $e_i$ points tangentially to it. Precisely: (note that $\mu_d = \alpha$)
\begin{equation}
\label{eqdef:CED}
\mu_i := \alpha + (1-\alpha) \exp\left(-\left(\frac \lambda {\lambda_d - \lambda_i}\right)^{m}\right).\\ %e^{-\left(\frac 
\end{equation}
The above formula often leads to false positives: at a position with large gradients, but without a clear preferred direction, one may very well have $\lambda_d - \lambda_i \gg \lambda$ (for instance if $\lambda_d = 100 \lambda$ and $\lambda_i = 95 \lambda$). A more reliable coherence detector is $\lambda_m - \lambda_i \gg \lambda + \lambda_i$, which leads to a Conservative variant of CED (cCED), see Figure \ref{fig:Triangle} for a comparison. Precisely: (note that $\mu_d=\alpha$)
\begin{equation}
\label{eqdef:cCED}
\mu_i := \alpha + (1-\alpha) \exp\left(-\left(\frac {\lambda+\lambda_i} {\lambda_d - \lambda_i}\right)^{m}\right).\\ %e^{-\left(\frac 
\end{equation}
\end{itemize}

We emphasize that the distinction between EED and its conservative variant cEED (resp.\ CED and cCED) is rather subtle, and mostly located around image contour corners, as evidenced on Figure \ref{fig:Triangle}. In other illustrations, we only show the conservative variant, which is slightly better at preserving detail.

\section{Implemented filters and their parameters}
\label{sec:ImplementationParameters}

\begin{figure} 
\includegraphics[width=7.5cm]{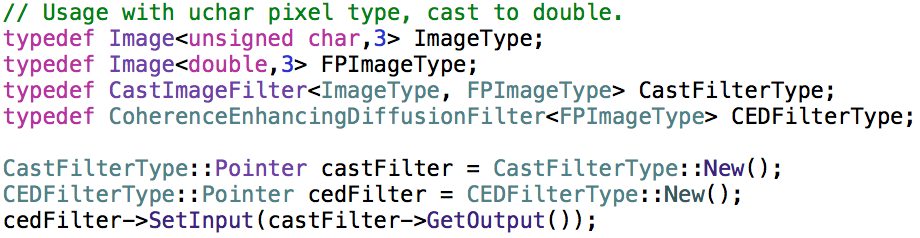}
\includegraphics[width=8.5cm]{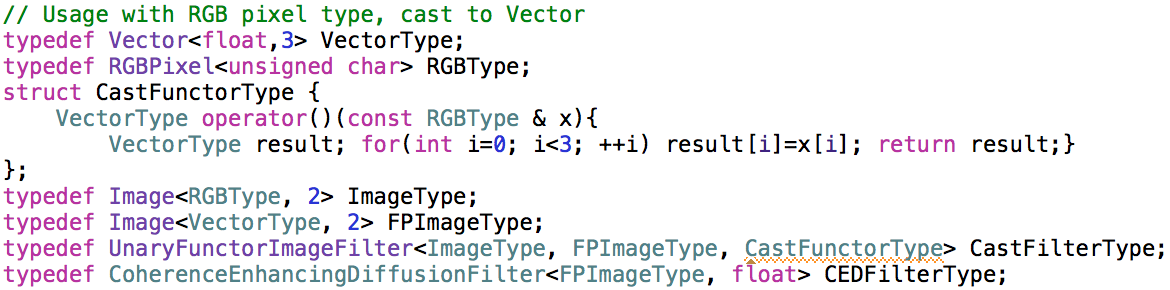}
\caption{The provided filters expect image pixels of scalar or vector pixel type. A preliminary cast is required for integral pixel types (left), or non-Vector pixel types such as RGB (right). The provided executable (or ``.cxx'' file) performs these casts automatically.
}
\label{fig:Cast}
\end{figure}

Our contribution to ITK consists of the following four image filters, which implement the mathematical notions presented in \S \ref{sec:LAD} and \S \ref{sec:NLAD}. The figures are produced with the last filter, which implements CED, EED and their variants described \S \ref{sec:NLAD}. The first three filters are its building blocks, but they may be of independent interest for other applications. All filters are multithreaded. 

For each filter, the processed image pixels can be of scalar or vector type. In the latter case, the underlying floating point type needs to be specified via the second template parameter of the filter, see Figure \ref{fig:Cast} (bottom right). Pixels of integral type (resp.\ RGB pixels) must be cast to floating point (resp.\ \doxygen{Vector}) types, see Figure \ref{fig:Cast}.  The image dimension must be $2$ or $3$.

\begin{description}
\item[Linear Anisotropic Diffusion (LAD).]  The filter \doxygen{LinearAnisotropicDiffusionLBRImageFilter}, requires two inputs: a processed image and a tensor image. Note that non-linear diffusion is achieved through successive linear diffusions, over multiple small time intervals, with regularly updated diffusion tensors.  Parameters:
\begin{itemize}
\item 
\emph{MaxDiffusionTime} specifies the target physical time for the LAD evolution PDE \eqref{eq:LinearAnisotropicDiffusion}. The filter has early abort options, as discussed in the next point, hence one should check the \emph{EffectiveDiffusionTime} at termination.
\item 
\emph{MaxNumberOfTimeSteps}, \emph{RatioToMaxStableTimeStep}. Explicit numerical schemes for diffusion are subject to a Courant-Friedrichs-Levy (CFL) condition, which limits the largest stable time step. 
This time step, which depends on the input diffusion tensors, is automatically computed by the filter, and the requested time interval $[0, \text{\emph{MaxDiffusionTime}}]$ is split accordingly. Early abort occurs if this splitting exceeds the specified \emph{MaxNumberOfTimeSteps}. 
%One may want to check the \emph{EffectiveNumberOfTimeSteps} at termination.
\end{itemize}

\item[Structure tensor.] Filter \doxygen{StructureTensorImageFilter}. Parameters:
\begin{itemize}
\item \emph{NoiseScale} $\sigma$, and \emph{FeatureScale} $\rho$, see \eqref{eq:StructureTensor}. Suggested ranges: $\sigma \in [0.5,3]$, $\lambda \in [2,10]$, assuming a unit pixel spacing. The lower bounds of these intervals are recommended, unless noise is extremely strong.
\item \emph{RescaleForUnitMaximumTrace}. If on, the structure tensors are rescaled: $S'_u = \alpha S_u$, where $\alpha>0$ is the largest constant such that $ {\rm Trace}( \Struct'_u(x) )\leq 1$ for all $x \in \Omega$. (The trace is the sum of the eigenvalues $\lambda_1+ \cdots + \lambda_d$, see \eqref{eq:StructToDiff}.) This option is meant to ease the choice of the edge detection threshold $\lambda$ in CED, EED. One may want to check the variable \emph{PostRescaling} $= \alpha$ at termination.

\end{itemize}

\item[Non-Linear Anisotropic Diffusion (NLAD).]
Filter \doxygen{AnisotropicDiffusionLBRImageFilter}.
This class (I) computes structure tensors by invoking the previous filter, (II) performs their eigen-analysis, (III) changes them into diffusion tensors via \eqref{eq:StructToDiff}, (IV) runs linear diffusion with the constructed tensors by invoking the first filter.
After a limited number of time-steps of linear diffusion, the steps (I-II-III-IV) are repeated so as to update the diffusion tensors, until exhaustion of the prescribed diffusion time. Parameters:
\begin{itemize}
\item \emph{DiffusionTime} for which the evolution rule \eqref{eq:NLAD} is applied.
\item \emph{Adimensionize.} If on, the filter ignores the image pixel spacing information, and sets on the \emph{RescaleForMaximumUnitTrace} option for structure tensor generation. This is intended to ease the choice of \emph{DiffusionTime}, and of the edge detection threshold $\lambda$ in CED, EED, and variants.
\item \emph{MaxTimeStepsBetweenTensorUpdates} is self descriptive. \emph{NoiseScale} and \emph{FeatureScale} are passed for structure tensor generation.
\item \emph{EigenValuesTransform} is a virtual method used to construct the diffusion tensor eigenvalues $(\mu_i)_{i = i}^d$ from those of the structure tensors $(\lambda_i)_{i=1}^d$, which are sorted increasingly for convenience.
The method must be redefined in a subclass, as in the next filter, else it triggers an exception.
% The class must be subclassed, and this method redefined, as in the next filter.
\end{itemize}

\item[Coherence-Enhancing diffusion and Edge-Enhancing diffusion.]
The two PDEs, and their variants, are implemented in the same filter \doxygen{CoherenceEnhancingDiffusionFilter}, which subclasses the Non-Linear Anisotropic Diffusion filter. Parameters:
\begin{itemize}
\item \emph{Enhancement} allows to switch between EED, cEED, CED, cCED and Isotropic \eqref{eqdef:Isotropic} tensor constructions, by redefining the superclass virtual method \emph{EigenValuesTransform}. The relevant choice depends on the type of image structures that one wants to enhance.
\item \emph{Lambda}$=\lambda$, \emph{Exponent}$=m$, \emph{Alpha}$=\alpha$ are the parameters involved in tensor design \eqref{eqdef:EED} - \eqref{eqdef:cCED}.
\item Suggested parameter ranges. \emph{Adimensionize}$=$True. \emph{DiffusionTime} $\in [0.5,\ 5]$, although larger values can be relevant for very strong noise or artistic effects. Edge detection threshold $\lambda \in [10^{-3}, \ 5 \times 10^{-2}]$, small for complex images with detail, large for simple ``cartoon'' like images. Finally $m \in [2,4]$, $\alpha=0.01$, though these parameters are secondary and have little impact. 
\end{itemize}

\end{description}

%\section{Application Examples}
%\label{sec:Examples}

%\paragraph{Aknowledgement:} The author thanks Jérôme Fehrenbach, Laurent Risser and Shaza Tobji for code review, and comparisons with their implementation of the same algorithm.

%%%%%%%%%%%%%%%%%%%%%%%%%%%%%%%%%%%%%%%%%
%
%  Insert the bibliography using BibTeX
%
%%%%%%%%%%%%%%%%%%%%%%%%%%%%%%%%%%%%%%%%%

\bibliographystyle{plain}
\iftoggle{arxiv}{
\bibliography{AllPapers}
}{
\bibliography{../../../AllPapers}
}

\end{document}